\documentclass{article}
\usepackage{ijcai21}

\usepackage{times}
\usepackage{soul}
\usepackage{url}
\usepackage[hidelinks]{hyperref}
\usepackage[utf8]{inputenc}
\usepackage[small]{caption}
\usepackage{graphicx}
\usepackage{amsmath}
\usepackage{amsthm}
\usepackage{booktabs}
\usepackage{algorithm}
\usepackage{algorithmic}
\usepackage{amsfonts,amssymb}
\title{Exploiting Spiking Dynamics with Spatial-temporal Feature Normalization in Graph Learning}

\author{
Mingkun Xu$^{\ast}$ 
\and
Yujie Wu\footnote{Contribute equally,   $^\dagger$Corresponding author}
\and
Lei Deng
\and
Faqiang Liu
\and
Guoqi Li \And
Jing Pei$^{\dagger}$   
\affiliations
Center for Brain-Inspired Computing Research (CBICR), Beijing Innovation Center \\for Future Chip, Department of Precision Instrument,  Tsinghua University\\
\emails
\{xmk18,wu-yj16,lfq18\}@mails.tsinghua.edu.cn,
\{leideng,liguoqi,peij\}@mail.tsinghua.edu.cn
}

\begin{document}

\maketitle

\begin{abstract}

Biological spiking neurons with intrinsic dynamics underlie the powerful representation and learning capabilities of the brain for processing multimodal information in complex environments. Despite recent tremendous progress in spiking neural networks (SNNs) for handling Euclidean-space tasks, it still remains challenging to exploit SNNs in processing non-Euclidean-space data represented by graph data, mainly due to the lack of effective modeling framework and useful training techniques. Here we present a general spike-based modeling framework that enables the direct training of SNNs for graph learning. Through spatial-temporal unfolding for spiking data flows of node features, we incorporate graph convolution filters into spiking dynamics and formalize a synergistic learning paradigm. Considering the unique features of spike representation and spiking dynamics, we propose a spatial-temporal feature normalization (STFN) technique suitable for SNN to accelerate convergence. We instantiate our methods into two spiking graph models, including graph convolution SNNs and graph attention SNNs, and validate their performance on three node-classification benchmarks, including Cora, Citeseer, and Pubmed. Our model can achieve comparable performance with the state-of-the-art graph neural network (GNN) models with much lower computation costs, demonstrating great benefits for the execution on neuromorphic hardware and prompting neuromorphic applications in graphical scenarios.
\end{abstract}


\section{Introduction}
Spiking neural network (SNN) is a typical brain-inspired network, which inherits the biological spatio-temporal dynamics mechanisms and rich spiking coding schemes, demonstrating great benefits for mimicking neuroscience-inspired models and performing efficient computation in neuromorphic hardware. Recently SNNs have been successfully applied in a variety of applications, such as image classification, voice recognition, object tracking, neuromorphic perception, etc \cite{wu2018spatio,silva2017evolving,cao2015spiking,zhu2020retina,haessig2019spiking}. However, most of published works have mainly focused on processing unstructured information represented in the Euclidean space, such as image, language data, etc \cite{zhou2018graph}. On the other hand, biological spiking neurons with rich coding schemes and spatio-temporal dynamics underlie the foundation of graphic cognitive map in the hippocampal-entorhinal system. It facilitates the organization and assimilation of internal association attributes in structured spatial representations for performing relational memory and advanced cognition \cite{whittington2020tolman}. Hence, it is highly expected to explore an effective algorithm for training SNNs involved with topology relations and structural knowledge.

Recently, there exists some works of combining SNNs with graphic scenarios, whereas these studies mainly focused on applying graph theory to analyze the features of spiking neuron and network topology \cite{piekniewski2007emergence,cancan2019ev,jovanovic2016interplay}, or using the features of spiking neurons to solve simple graph-related problems, such as the shortest path problems, clustering problems, minimal spanning tree problems \cite{sala1999solving,hamilton2020spike}. More recently, the work \cite{gu2020tactilesgnet} introduced a graph convolution to pre-process tactile data and trained an SNN classifier. Although they achieved high performance on the classification of sensor data, this method is difficult to accommodate general graph operations and transfer to other scenarios. On the other hand, plenty of GNN models have been developed, such as Graph Convolution Network (GCN) and Graph Attention Network (GAT), and obtained remarkable progress in solving graph-related tasks \cite{kipf2019semi,velivckovic2017graph}. While few works considered the relationship between graph theory and neural dynamics, and most of them suffered from a huge computational overhead when applied in scenarios with large-scale data, potentially limiting practical applications.

To explore the feasibility and advantages of spiking dynamics for handling graph-structured data, here we report a graph SNN framework capable of processing non-Euclidean data with general operations by introducing versatile aggregation methods. To develop such SNN models, two main challenges need to overcome. The first is to reconcile multiple graph convolution operations and spiking dynamics. To this end, we firstly unfold the data flows of binary node features along the temporal dimension and spatial dimension, and propose a general spiking message passing method incorporating graph filter and spiking dynamics in an iterative manner. The second is to ensure the training convergence and performance for graph SNNs. Because of the complex spiking dynamics and diverse graph operations, directly training SNNs on graph is still rarely researched and challenging. Although various normalization techniques have demonstrated their powerful effectiveness for the network convergence \cite{ioffe2015batch,ba2016layer}, it is difficult to apply these techniques directly to graph SNNs especially owning to temporal neuronal dynamics and binary communication mechanism \cite{zheng2020going}. Therefore, it is necessary to develop a new normalization method suitable for spiking dynamics and graph scenarios.  To address this problem, we propose a spatial-temporal feature normalization algorithm (STFN) by normalizing instant membrane potentials corresponding to each node in feature dimension and temporal dimension. In this manner, SNNs can gain a powerful capacity for abstracting hidden features from aggregated signals in a graph. Overall, our contributions are summarized as follows:
\begin{itemize}
\item  We build a general spike-based modeling framework, referred to as Graph SNNs, to support spike propagation and feature affine transformation by reconciling the graph convolution operation and spiking communication mechanism. The framework is transferable and can accommodate most of graph propagation operations. To our best knowledge, this is the first work to build a general spike-based framework for processing regular graph tasks in gradient-descent paradigm.
\item  We propose a spatial-temporal feature normalization algorithm (STFN), which can incorporate the temporal neuronal dynamics and coordinate the membrane potential representation with threshold, demonstrating a significant improvement on convergence and performance.
\item  We instantiate the proposed framework into several models (GC-SNN and GA-SNN) and demonstrate their performance by multiple experiments on the semi-supervised node classification tasks. We further evaluate the computation costs and demonstrate high-efficiency benefits which may bring new opportunity to unlock SNNs potentials and facilitate graph-structured applications for neuromorphic hardware.
\end{itemize}

\begin{figure*}[h]
\begin{center}
\includegraphics[height=9.0cm,width=14.5cm]{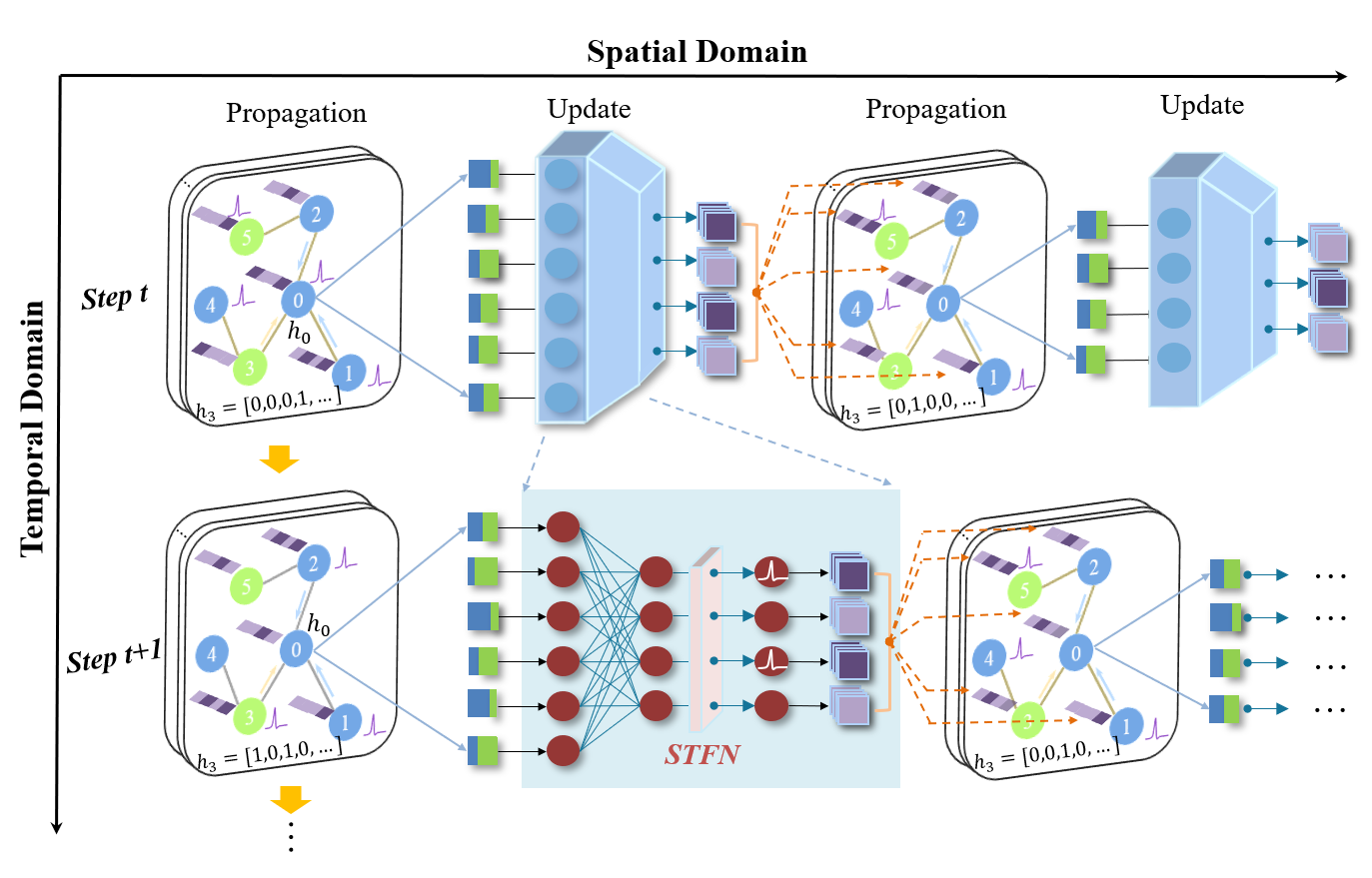}
\end{center}
\caption{The Graph SNN framework can support spiking message propagation and feature affine transformation, reconciling the graph convolution operation and spiking communication mechanism in a unified paradigm. The proposed STFN normalizes the membrane potentials along both spatial and temporal dimension, which coordinates the data distribution with threshold but also facilitates the network convergence.
}
\label{architecture}
\end{figure*}

\section{Graph Spiking Neural Networks}
We introduce our Graph SNN framework mainly from following four aspects: (1) an approach to accommodate graph convolution operation into spiking computing paradigm; (2) an iterative spiking message passing model and the overall training framework containing two phases; (3) the spatial-temporal feature normalization for facilitate convergence; (4) instantiations of specific spiking graph model.

\subsection{Spiking Graph Convolution}
Given an attributed graph $ \mathcal{G}= (\mathcal{V}, \mathcal{E})$,  $\mathcal{V}$ is the set of nodes, and $\mathcal{E}$ is the set of edges. Generally, the graph attributes can be described by an adjacent matrix $A \in \mathbb{R}^{N \times N} $ and a node feature matrix $X \in \mathbb{R}^{N \times C}=[x_1;x_2;...;x_N]$, where each row of $X$ represents the feature vector of each node $x_i$ and the adjacent matrix $A$ satisfies $A_{ii}=0$. Assume the single node signal is represented as a scalar vector $x \in \mathbb{R}^C$, the spectral convolution is defined as the multiplication of a filter $\mathcal{F}_\theta = diag(\theta) $ and node signal $x$ as below:
\begin{equation}
 \mathcal{F}_\theta \star x = U \mathcal{F}_\theta U^{\top} x,
\end{equation}
where $\mathcal{F}_\theta $ is parameterized by $\theta \in \mathbb{R}^C$, $U$ is the eigenvector matrix of Laplacian matrix $L=I - D^{-\frac{1}{2}}AD^{-\frac{1}{2}}=U \Lambda U^{\top}$. $ U^{\top} x$ can be regarded as the graph Fourier transform of $x$, and the diagonal matrix of eigenvalues $\Lambda$ can be filtered by the function $\mathcal{F}_\theta(\Lambda)$.  Due to the expensive computation overhead, some work proposed to use  approximation and stacked multiple nonlinear layers to circumvent it, which has been applied successfully recently \cite{kipf2019semi}.

To unify the information flows in spiking dynamics, we encode initial node signals $X$ into binary components $ \{\widetilde{X}_0, \widetilde{X}_1, ...,\widetilde{X}_t,..., \widetilde{X}_{T-1} \} $, where $T$ signifies the length of time window. The encoding method can be probabilistic, such as following a Bernoulli distribution or Poisson distribution, and can also be deterministic, such as quantitative approximation or using an encoding layer to generate spikes \cite{wu2019direct}. In this manner, we can transform an attributed graph from its initial state into a spiking expression. We use $\widetilde{H}_t^n$ to represent the spiking node embedding in the $n_{th}$ layer at time $t$, and $\widetilde{H}_t^0 = \widetilde{X}_t$ (tilde denotes binary variable in spike form). Then the layer-wise spiking graph convolution in spatial-temporal domain can be defined as:
\begin{equation}
\widetilde{H}_t^{n} =  \Phi (\mathcal{G}_c(A, \widetilde{H}_t^{n-1}) W^n, \widetilde{H}_{t-1}^{n}),
\label{spi-conv}
\end{equation}
where $\mathcal{G}_c(A, \widetilde{H})$ denotes the spiking feature propagation on the topology connection structure,  which can be implemented via various propagation methods, such as Chebyshev filter, $1^{st}$ model, etc. $\Phi(\cdot)$ is the non-linear dynamic process related to historical states $\widetilde{H}_{t-1}^{n}$. $W^n \in \mathbb{R}^{C^{n-1} \times C^n}$ is the layer-wise trainable weight parameter matrix, where $C^n$ denotes the output dimension of spiking feature in the $n_{th}$ layer and $C^0=C$. We use the formula (\ref{spi-conv}) to represent one spiking graph convolution layer. And multi-layer-stacked structure can form a comprehensive spiking dynamics system.

\subsection{Iterative Spiking Message Passing}
We adopt the leaky integrate-and-fire (LIF) model as our basic neuron unit, which is computationally tractable and most commonly used, meanwhile maintaining biological fidelity to a certain degree. Its neuronal dynamics can be governed by:
\begin{equation}
\label{lif}
\tau \frac{dV(t)}{dt} = -(V(t)-V_{reset}) + I(t),
\end{equation}
where $V(t)$ denotes the neuronal membrane potential at time t. $\tau$ is a time constant and $I(t)$ signifies the pre-synaptic input, which is obtained by the combined action of synaptic weights and pre-neuronal activities or external injecting stimulus. When $V(t)$ exceeds a certain threshold $V_{th}$, the neuron emits a spike and resets its membrane potential to $V_{reset}$, then starts to accumulate $V(t)$ again in subsequent time steps.

The spiking message passing process includes information propagation step and update step, both of which run for $T$ time steps. Let $\widetilde{H}_t=[\widetilde{h}_t^0;\widetilde{h}_t^1;...;\widetilde{h}_t^{N-1}] \in \mathbb{R}^{N \times C}$, for the node $v$, we provide a universal formulation for its spiking message passing as:
\begin{equation}
\label{mpss}
\widetilde{h}^v_t = U(\sum_{u \in N(v)}P(\widetilde{h}^v_t, \widetilde{h}^u_t, e_{vu}),\widetilde{h}^v_{t-1})),
\end{equation}
where $P(\cdot)$ represents the spiking message aggregation for neighbouring nodes, which can be implemented by different graph convolution operation $\mathcal{G}_c(\cdot)$. $U(\cdot)$ represents the state update via a non-linear dynamic system. $N(v)$ denotes all the neighbors of node $v$ in graph $\mathcal{G}$, and $e_{vu}$ denotes the static edge connection between $v$ and $u$, which can be extended to directed multigraph formalism trivially. Equation $(\ref{spi-conv})$ is a specific implementation of the above universal expression.

To accommodate the LIF model into the above paradigm, we use Euler method to transform the first-order differential equation of Eq.$(\ref{lif})$ into an iterative expression.
We use a decay factor $\kappa$ to represent the item ($1 - \frac{dt}{\tau}$) and expand the pre-synaptic input $I$ as $\sum_j W^j \mathcal{G}_c(A, \widetilde{H}^j_{t+1})$. Note that here we use graph convolution to realize propagation step $P(\cdot)$, then we obtain the following formulation via incorporating the scaling effect of $\frac{dt}{\tau}$ into weight item:
\begin{equation}
V_{t+1} = \kappa V_t + \sum_j W^j\mathcal{G}_c(A, \widetilde{H}^j_{t+1}).
\end{equation}
$\mathcal{G}_c(A, \widetilde{H}^j_{t+1})$ denotes the corresponding pre-synaptic aggregated feature and the superscript $j$ denotes pre-synapse index. By incorporating firing-and-resetting mechanism and assuming $V_{reset}=0$, the update equation can be formulated as:
\begin{align} 
&  V^{n,i}_{t+1} = \kappa V^{n,i}_t(1-\widetilde{H}^{n,i}_t) + \sum^{l(n)}_j W^{n,ij} \mathcal{G}_c(A, \widetilde{H}^{n-1,j}_{t+1}), \label{update1} \\
& \widetilde{H}^{n,i}_{t+1} = g(V^{n,i}_{t+1} - V_{th}), \label{firing} 
\end{align} 
where $n$ denotes the $n_{th}$ layer and $l(n)$ denotes its corresponding neuron number. $W^{ij}$ represents synaptic weight from $j_{th}$ neuron in pre-layer to $i_{th}$ in post-layer. $g(\cdot)$ is Heaviside step function. In this manner, we convert the implicit differential function into an explicit iterative version, which can describe the message propagation and update process in formula $(\ref{mpss})$ and Figure $1$. 

Additionally, we find that the computation order of affine transformation and graph convolution can be exchanged when the graph convolution operation is linear (e.g. $\mathcal{G}_c(A, H) = D^{-\frac{1}{2}} AD^{-\frac{1}{2}}H $). Considering that $W$ is quite dense,  $\widetilde{H}$ is very sparse and binary, the prioritizing calculation for $\widetilde{H}$ and $W$ will reduce computation overhead by turning multiplication into addition. In this situation, the process $(\ref{update1})$ can be re-formulated as:
\begin{equation}
\label{update2}
V^{n,i}_{t+1} = \kappa V^{n,i}_t(1-\widetilde{H}^{n,i}_t) + \mathcal{G}_c(A, \sum^{l(n)}_j W^{n,ij} \widetilde{H}^{n-1,j}_{t+1}).
\end{equation}
In this manner, we provide a universal spiking passing framework in iterative manner. By specializing $\mathcal{G}_c(\cdot)$,  most of proposed graph convolution operations \cite{kipf2019semi,gilmer2017neural,hamilton2017inductive} can be incorporated into this compatible model. And this framework is transferable to other graph scenarios.

\subsection{Spatial-temporal Feature Normalization}
Due to the additional temporal dynamics and event-driven spiking binary representation, traditional normalization techniques cannot be applied to the SNN framework directly. Moreover, on the other hand, the convergence and performance cannot be guaranteed by directly training SNNs on graph tasks. Therefore, it motivates us to propose a spatial-temporal feature normalization (STFN) algorithm specialized for the spiking dynamics on graph scenarios.

Considering the feature map calculation step, let $S_t \in \mathbb{R}^{N \times C}$ represent the instant outputting membrane potential of all neurons in a layer at time step $t$ ($\sum^{l(n)}_j W^{n,ij} \mathcal{G}_c(A, \widetilde{H}^{n-1,j}_{t})$ in formula $(\ref{update1})$ or $\sum^{l(n)}_j W^{n,ij} \widetilde{H}^{n-1,j}_{t}$ in formula $(\ref{update2})$). 
In STFN, the pre-synaptic inputs will be normalized along the feature dimension $C$ independently for each node. Considering that the temporal effects play a significant role in operating node features transformation in topology space, we simultaneously normalize along the temporal dimension at consecutive time steps. Let $S^{k,v}_t$ represent the $k_{th}$ element in feature vector of a node $v$. $S^{k,v}_t$ will be normalized by
\begin{equation}
\label{stfn}
\begin{aligned} 
& \hat{S}_t^{k,v} = \frac{\rho V_{th}(S_t^{k,v}-E[S^v])}{\sqrt{Var[S^v] + \epsilon}},
\\
& Y_t^{k,v} = \lambda^{k,v} \hat{S}_t^{k,v} + \gamma^{k,v},
\end{aligned} 
\end{equation}
where $\rho$ is a hyper-parameter optimized in the training process and $\epsilon$ is a tiny constant. $\lambda^{k,v}$ and $\gamma^{k,v}$ are two trainable parameter vectors for node $v$. $E[S^v]$ and $Var[S^v]$ denote the mean and variance of node $v$ statistically computed over feature dimension and temporal dimension, respectively. Figure $2$ depicts the computation process of  $E[S^v]$ and $Var[S^v]$, which are defined as follows:
\begin{equation}
\label{mean-var}
\begin{aligned} 
& E[S^v] = \frac{1}{CT}\sum^{T-1}_{t=0} \sum^{C-1}_{k=0} S^{k,v}_t,
\\
& Var[S^v] =  \frac{1}{CT}\sum^{T-1}_{t=0}\sum^{C-1}_{k=0} (S^{k,v}_t - E[S^v])^2.
\end{aligned} 
\end{equation}

\begin{figure}[h]
\begin{center}
\includegraphics[height=5.0cm,width=8.5cm]{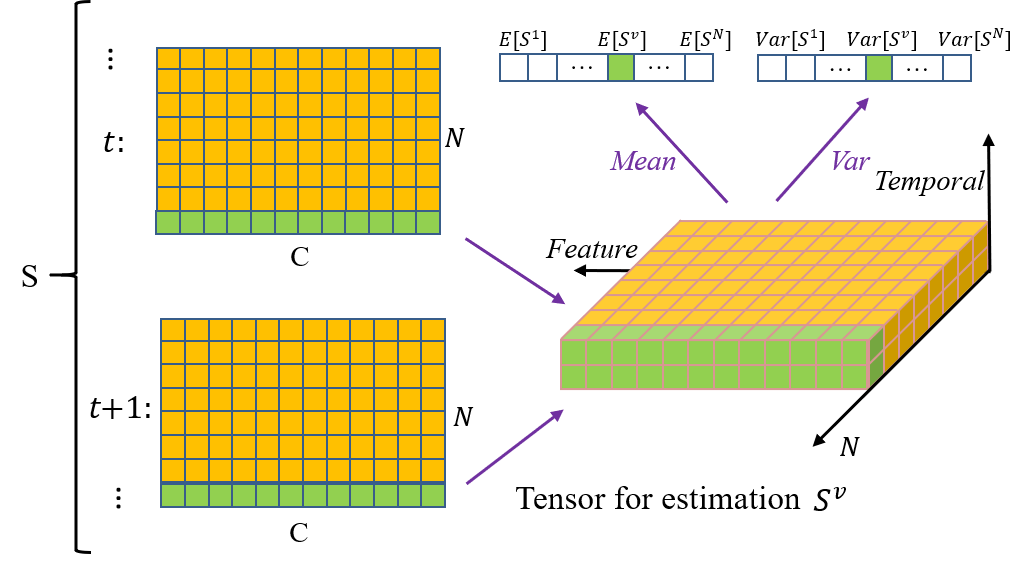}  %
\end{center}
\caption{The schematic diagram of spatial-temporal feature normalization (STFN), where the high-dimensional pre-synapse inputs will be normalized along the feature dimension and temporal dimension. 
}
\label{architecture}
\end{figure}
We follow the above schema and find the normalization can facilitate convergence, improve performance in training. 

\subsection{Graph SNNs Framework}
Our framework includes two phases: spiking message passing and readout, wherein the spiking message passing phase is described in Section $2.2$. After the  $T$ times of iteration for the inference process are finished, the output spiking signals need to be decoded and transformed into high-level representations for downstream tasks. Under the rate coding condition, the readout phase computes the decoded feature vector for the whole graph using readout function $R$ according to
\begin{equation}
\label{readout}
\begin{aligned} 
\hat{y}^v = R(\{\frac{1}{T}\sum^{T-1}_{t=0}\widetilde{h}^v_t | v \in \mathcal{G} \}).
\end{aligned} 
\end{equation}
Here, the readout function operates on the set of node states and should be invariant to permutations of the node states in order that Graph SNNs can be invariant to graph isomorphism. Besides, the readout function can be implemented via multiple approaches, such as using a differentiable function or neural network that maps to an output.

As shown in Figure $1$, the first phase of our framework contains two unfolding domains, i.e., spatial domain and temporal domain. In the temporal domain, we utilize the encoded binary node spikes as the current node feature at time step $t$. A sequence of binary information will be processed successively, where the historical information for each neuron plays a key role in spiking dynamics and the decoding for output. In the spatial domain, the spiking features of a batch of nodes will be firstly aggregated via connected edges from one-hop neighbors, where the propagation operation is compatible with multiple proposed methods. Then the aggregated features will be fed into the spiking network module. In each network module, we conduct STFN to normalize the instant outputting membrane potential along the feature and temporal dimension, which ensures the node features of all dimensions follow the distribution $N(0,(\rho V_{th})^2)$. Each layer possesses the spatial-temporal dynamics for feature abstracting and mapping, and the stacked multi-layer structure will empower SNNs stronger representation capability. The inference procedure is provided in Supplementary Materials.

We take the semi-supervised learning task as example for illustration. With a small subset of node labels available, we evaluate the cross-entropy error over all labeled examples:
\begin{equation}
\label{loss}
\begin{aligned} 
\mathcal{L} = - \sum_{l\in \mathcal{Y}_L} \sum^{C^L-1}_{r=0} Y_{lr}ln \sigma(\hat{y}_{lr}), 
\end{aligned} 
\end{equation}
where $\sigma(\cdot)$ denotes softmax function, $\mathcal{Y}_L$ represents the set of node indices possessing labels. $Y_{lr}$ denotes the real labels corresponding to the $r_{th}$ dimension of $C^L$ classes. In order to train the Graph SNN models effectively via gradient descent paradigm, we adopt the gradient substitution method in backward path \cite{wu2018spatio} and take the rectangular function to approximate the derivative of spike activity.

\subsection{Model Instantiating: GC-SNN and GA-SNN}
To demonstrate the effectiveness of our framework and normalization technique, we instantiate the framework into specific models by introducing specifically designed propagation methods most used in GNNs: graph convolution aggregator and graph attention mechanism \cite{kipf2019semi,velivckovic2017graph}. In this case, our Graph convolution Spiking Neural Network (GC-SNN) can be formulated as:

\begin{equation}
\widetilde{h}_t^{n,i} =  \Phi (\sum_{j\in \mathcal{N}(i)}\frac{1}{c_{ij}} \widetilde{h}_t^{n-1,j}W^n+ b^n, \widetilde{h}_{t-1}^{n}),
\label{gc-snn}
\end{equation}
where $\mathcal{N}(i)$ is the set of neighbors of node $i$, $c_{ij}$ is the product of square root of node degrees (i.e. $c_{ij}=\sqrt{\mid \mathcal{N}(i)\mid}\sqrt{\mid \mathcal{N}(j)\mid}$). $b^n$ is a trainable bias parameter. Similarly, our Graph Attention Spiking Neural Network (GA-SNN) can be expressed as:
\begin{equation}
\begin{aligned} 
& \widetilde{h}_t^{n} =  \Phi (\alpha_{ij}^n \sum_{j \in \mathcal{N}(i)}\widetilde{h}_t^{n-1,j} W^n, \widetilde{h}_{t-1}^{n}),  \\ 
& \alpha_{ij}^n = \frac{exp(f_l({a^n}^T(\widetilde{h}_t^{n-1,i} W^n || \widetilde{h}_t^{n-1,j} W^n)))}{\sum_{k \in \mathcal{N}(i)}exp(f_l({a^n}^T(\widetilde{h}_t^{n-1,i} W^n || \widetilde{h}_t^{n-1,k} W^n)))},  \\
\end{aligned} 
\label{gc-snn}
\end{equation}
where $f_l(\cdot)$ denotes $LeakyReLu(\cdot)$ function. $a^n$ is a learnable weight vector and $||$ denotes the concatenation operation. The core idea is to conduct attention mechanism on graph convolution and learn different importance degree $\alpha$ for each node. We exhibit the results of the above models for illustration in the experiment section.

\section{Experiments}
In this section, We firstly investigate the capability of our models over semi-supervised node classification on three citation datasets to examine their performance. Then we demonstrate the effectiveness of the proposed STFN and provide some analysis visualization for the model efficiency. We provide the  experimental configuration details in Supplementary Materials.

\subsection{Datasets and Pre-processing}
We use three standard citation network benchmark datasets---Cora, Pubmed, and Citeseer, where nodes represent paper documents and edges are (undirected) citation links. We summarize the dataset statistics in Supplementary Materials. Note that from bag-of-words model, the features can be represented as binary vectors, which is equipped with similar attributes with spike representation. Here, we take the binary representation as spike vector, and re-organized them as an assembling sequence set w.r.t timing, where each component vector is considered equal at different time step for simplicity. In this manner, the original signals can be processed and treated as spiking signals.


\subsection{Performance and Analysis}

\begin{table}
\centering
\begin{tabular}{lccc}
\toprule
Method  &  Cora  &  Pubmed  & Citeseer \\
\midrule
DeepWalk   & 67.2  & 65.3   &  43.2 \\
ICA     & 75.1  & 73.9   &  69.1 \\
Planetoid     & 75.7   & 77.2    &  64.7   \\
Chebyshev   & 81.2 & 74.4  &   69.8  \\
GCN   & 81.5 & 79.0  &   70.3  \\
\midrule
GCN$^\ast$   & 81.9$\pm$1.1 & 79.4$\pm$0.4  &  70.4$\pm$1.1  \\
GAT$^\ast$   & 82.3$\pm$0.6 & 78.4$\pm$0.5  &  71.1$\pm$0.2  \\
GC-SNN(Ours)   & \textbf{80.7}$\pm$0.6 & 77.9$\pm$0.5  &  \textbf{69.9}$\pm$0.9  \\
GA-SNN(Ours)    & 79.7$\pm$0.6 & \textbf{78.0}$\pm$0.4  &  69.1$\pm$0.5  \\
\bottomrule
\end{tabular}
\caption{
Performance comparison on benchmark datasets \protect\cite{perozzi2014deepwalk,getoor2005link,yang2016revisiting,defferrard2016convolutional}. $\ast$ denotes the results in our implementation and $\pm$ denotes the standard deviation calculated from $10$ runs.}
\label{performance}
\end{table}

\begin{table}
\centering
\begin{tabular}{lccc}
\toprule
Operations($\times10^6$)  &  Cora  &  Pubmed  & Citeseer \\
\midrule
GNN(2 layers)     & 2.78  & 7.82   &  3.02 \\
SNN(2 layers)     & 0.22  & 0.34   &  0.26 \\
GNN(3 layers)     & 143.24   & 1018.66    &  175.84   \\
SNN(3 layers)     & 3.63   & 24.36    &  3.28   \\
\midrule
Compre. ratio(2 layers) $^\ast$  & 12.62$\times$ & 23.00$\times$ & 11.62$\times$  \\ %
Compre. ratio(3 layers) $^\ast$  & 39.46$\times$ & 41.82$\times$  &  53.61$\times$ \\ 
\bottomrule
\end{tabular}
\caption{Operation comparison on benchmark datasets. $\ast$ denotes the compression ratio (GNN Opts. / SNN Opts.) in feature transformation process.}
\label{performance}
\end{table}

\begin{figure*}[h]
\begin{center}
\includegraphics[height=3.95cm,width=16.5cm]{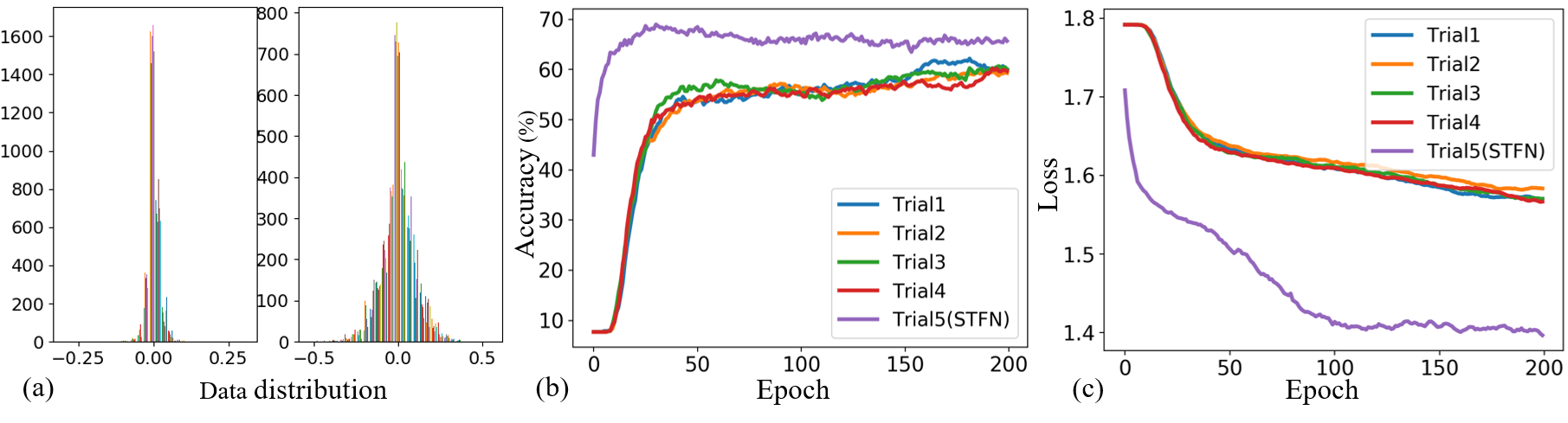}  
\end{center}
\caption{(a) The pre-activated membrane potential distribution before STFN (Left) and after STFN (Right). (b) Test accuracy comparison during training on Citeseer between Graph SNN models with STFN and without STFN. (c) Test loss comparison during training on Citeseer between Graph SNN models with STFN and without STFN.
}
\label{stfn-ana}
\end{figure*}

\begin{figure*}[h]
\begin{center}
\includegraphics[height=3.95cm,width=16.5cm]{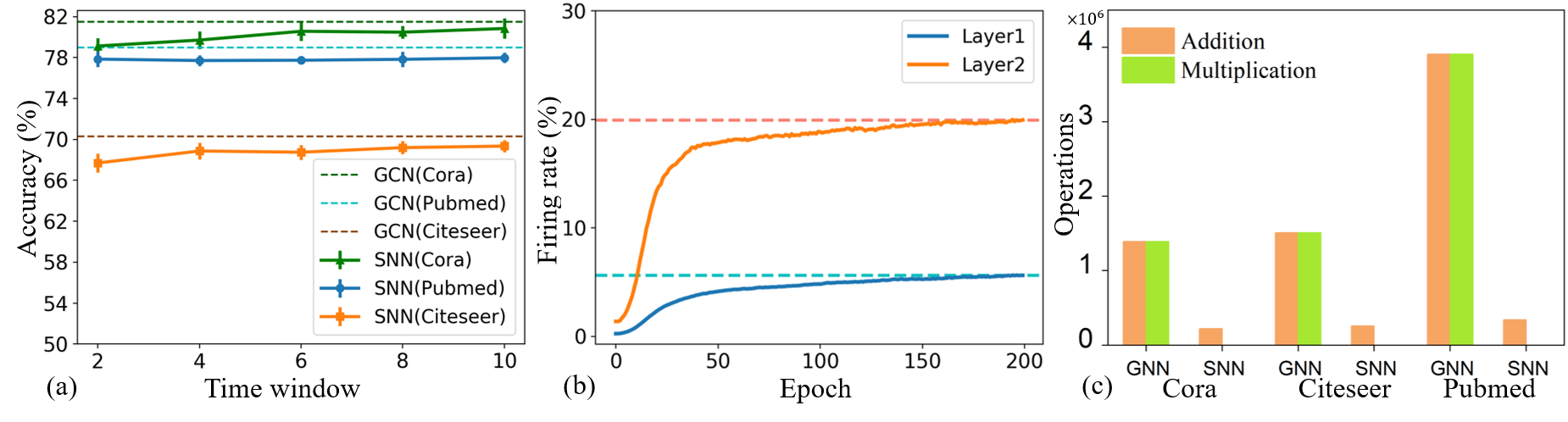}   
\end{center}
\caption{(a) Accuracy variation with respect to time window length (error bar denotes standard deviation from 10 runs). (b) Averaged firing rate variation during training process. (c) Operation cost comparison of feature transformation between GNNs and SNNs. 
}
\label{benefit}
\end{figure*}

\paragraph{Basic Performance.} For fairness, we take the same settings for GNN models (GCN, GAT) and our SNN models (GC-SNN, GA-SNN), and also keep the settings of each dataset same. We report our results for 10 trials in Table $1$. The results suggest that even if using binary spiking communication, our SNN models can achieve comparable performance with the SOTA results with a minor gap. It proves the feasibility and powerful capability of spiking mechanism and spatial-temporal dynamics, which can bind diversiform features from different nodes and work well on graph scenarios with few labels.

\paragraph{The Effects of STFN.} We further visualize the data distribution in Figure $3(a)$, which indicates that the STFN can push the feature distribution to follow a larger standard deviation controlled by the threshold. In this case, the normalized membrane potential state can coordinate distribution difference and spike-triggered threshold. We find that this method contributes to the stability of performance via multiple experiments. Additionally, we plot the comparison curves for test accuracy and loss during the training process(Figure $3(b)(c)$) on Citeseer. It also indicates that STFN can accelerate convergence, alleviate over-fitting, and significantly improve the performance on semi-supervised learning tasks.

\paragraph{The Effects of Time Window.} We conduct multiple experiments with different time window lengths. As depicted in Figure $4(a)$, a longer time window length will benefit for the accuracies  progressively, while the dependency on time window length is not prominent. Evidently, our model can still achieve comparable results even with a small time window and training cost. 

\subsection{Efficiency}
As presented in Figure $4(b)$, because we use the second layer for decoding and determining, the neuron activity density is much higher than that in the first layer. In the overall training process, the highest averaged firing rate is no more than $20\%$, and the overwhelming majority of neurons fire less than $10\%$ (see Appendix), verifying the prominent sparsity of spike features. More details are provided in Supplementary Materials.

We compare the affine transformation costs of GNN network and SNN network with 2-layer structure on three datasets, respectively. As shown in Figure $4(c)$, our SNN models have no addition operation and achieve far fewer multiplication operations than GNN models (compression ratio $11.62\times\sim23.00\times $). In Table $2$, the advantages become more prominent with deeper and larger network structure where the compression ratio become $39.46\times\sim53.61\times$. Fundamentally, the spike communication and even-driven computation underlie the efficiency advantage of SNNs. On the one hand, the binary spiking feature can transform the matrix manipulation with dense weights from multiplication into addition, where the energy consumption for addition is less than multiplication on neuromorphic hardware. On the other hand, the event-driven characteristics enable sparse firing and representation for node features, reducing the operation costs by a large margin. Besides, since the spike-based communication and local computation can be fully leveraged by neuromorphic hardware which adopts distributing memory and self-contained computational units in decentralized many-core architecture, our model provides a templet promoting the graphic application for neuromorphic computing.

\section{Conclusion}
In this work, we report a general SNN framework for graph-structured data with an iterative spiking message passing method. By proposing the STFN method, we incorporate the graph propagation and spiking dynamics into one unified paradigm and reconcile them in a collaborative mode. Our framework is flexible and transferable for various propagation operations and scenarios, and we instantiate it into two specific models for demonstrations: a GC-SNN for graph convolution and a GA-SNN for graph attention. The experimental results on three benchmark datasets demonstrate the model effectiveness and powerful representation capability. More importantly, Graph SNNs possess high-efficiency advantages conducive to the implementation on neuromorphic hardware and applications for graphic scenarios. Overall, this work sheds new light on spiking dynamics research interacted with graphic topology, which may facilitate the understanding for advanced cognitive intelligence.

\section*{Acknowledgments}
This work was supported by National Key R$\&$D Program of China (2018YFE0200200), National Nature Science Foundation of China (No61836004), Beijing Science and Technology Program (Z191100007519009) and CAAI-Huawei MindSpore Open Fund.

\bibliographystyle{named}
\bibliography{ijcai21}

\end{document}